\documentclass[11pt]{article}
\usepackage{authblk}
\usepackage{eacl2017}
\usepackage{times}
\usepackage{url}
\usepackage{latexsym}
\usepackage{blindtext}  
\usepackage{graphicx} 
\usepackage{amsmath} 
\usepackage{multirow}   
\usepackage{color}   
\usepackage{scalefnt} 
\usepackage{subfigure} 
\usepackage{float}  
\usepackage{tabularx} 
\usepackage[table,xcdraw]{xcolor} 
\usepackage[normalem]{ulem} 
\usepackage{mathptmx} 
\usepackage{anyfontsize} 
\usepackage{t1enc} 

\useunder{\uline}{\ul}{} 

\let\oldtabular\tabular 
\renewcommand{\tabular}{\footnotesize\oldtabular} 

\eaclfinalcopy 
\def\eaclpaperid{384} 

\title{Be Precise or Fuzzy:\\Learning the Meaning of Cardinals and Quantifiers from Vision}

\author[1]{\bf Sandro Pezzelle}
\author[2]{\bf Marco Marelli}
\author[3]{\bf Raffaella Bernardi}
\affil[1]{CIMeC - Center for Mind/Brain Sciences, University of Trento}
\affil[2]{Department of Psychology, University of Ghent}
\affil[3]{CIMeC/DISI, University of Trento}
\affil[  ]{\tt  \{sandro.pezzelle\}@unitn.it}
\affil[ ]{\tt  \{marco.marelli\}@ugent.be}
\affil[ ]{\tt  \{raffaella.bernardi\}@unitn.it}

\date{}

\begin{document}
\maketitle

\begin{abstract}
{\fontsize{10}{12}\selectfont People can refer to quantities in a visual scene by using either exact cardinals (e.g. \textit{one}, \textit{two}, \textit{three}) or natural language quantifiers (e.g. \textit{few}, \textit{most}, \textit{all}). In humans, these two processes underlie fairly different cognitive and neural mechanisms. Inspired by this evidence, the present study proposes two models for learning the objective meaning of cardinals and quantifiers from visual scenes containing multiple objects. We show that a model capitalizing on a `fuzzy' measure of similarity is effective for learning quantifiers, whereas the learning of exact cardinals is better accomplished when information about number is provided.}

\end{abstract}

\section{Introduction}
\label{sec:intro}

In everyday life, people can refer to quantities by using either cardinals (e.g. \textit{one}, \textit{two}, \textit{three}) or natural language quantifiers (e.g. \textit{few}, \textit{most}, \textit{all}). Although they share a number of syntactic, semantic and pragmatic properties~\cite{hurewitz2006}, and they are both learned in a fairly stable order of acquisition across languages~\cite{wynn1992,katsos2016}, these quantity expressions underlie fairly different cognitive and neural mechanisms. First, they are handled differently by the language acquisition system, with children recognizing their disparate characteristics since early development, even before becoming `full-counters'~\cite{hurewitz2006,sarnecka2004,barner2009}. Second, while the neural processing of cardinals relies on the brain region devoted to the representation of quantities, quantifiers rather elicit regions for general semantic processing~\cite{wei2014}. Intuitively, cardinals and quantifiers refer to quantities in a different way, with the former representing a mapping between a word and the exact cardinality of a set, the latter expressing a `fuzzy' numerical concept denoting set relations or proportions of sets~\cite{barner2009}. As a consequence, speakers can reliably answer questions involving quantifiers even in contexts that preclude counting~\cite{pietroski2009}, as well as children lacking exact cardinality concepts can understand and appropriately use quantifiers in grounded contexts~\cite{halberda2008,barner2009}. That is, knowledge about (large) precise numbers is neither necessary nor sufficient for learning the meaning of quantifiers.

\begin{figure}[t!]
\begin{center}
  \includegraphics[width=0.8\linewidth]{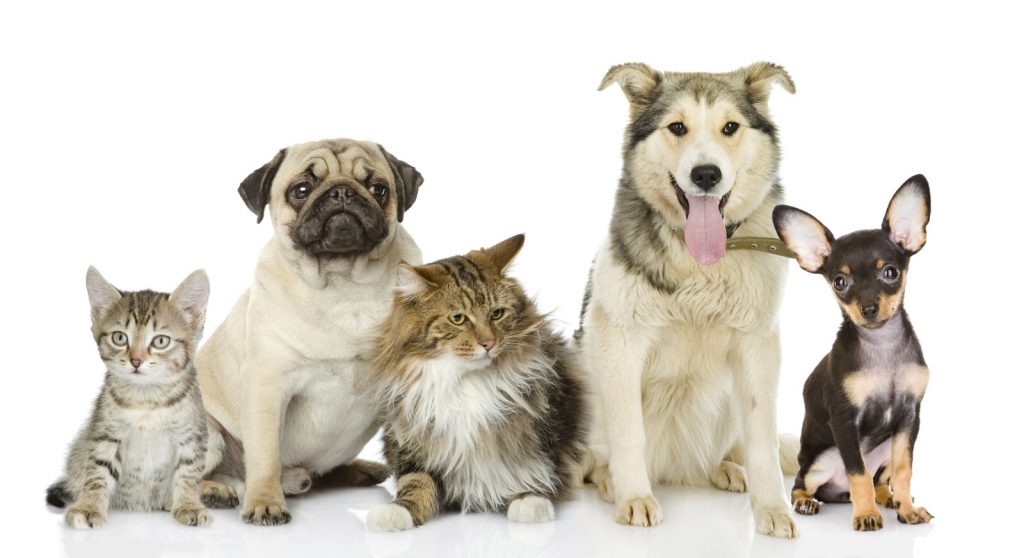}
\end{center}
\caption{How many are \textit{dogs}? Three/Most.}\label{fig:scenario}
\end{figure}

Inspired by this evidence, the present study proposes two computational models for learning the meaning of cardinals and quantifiers from visual scenes. Our hypothesis is that learning cardinals requires taking into account the number of instances of the target object in the scene (e.g. number of \textit{dogs} in Figure~\ref{fig:scenario}). Learning quantifiers, instead, would be better accomplished by a model capitalizing on a measure evaluating the `fuzzy' amount of target objects in the scene (e.g. proportion of `dogness' in Figure~\ref{fig:scenario}). In particular, we focus on those cases where both quantification strategies might be used, namely scenes containing target (\textit{dogs}) and distractor objects (\textit{cats}). Our approach is thus different from salient objects detection, where the distinction targets/distractors is missing~\cite{borji2015,sos2015,sos2016}. With respect to cardinals, our approach is similar to~\cite{segui2015}, who propose a model for counting people in natural scenes, and to more recent work aimed at counting either everyday objects in natural images~\cite{chatto2016} or geometrical objects with attributes in synthetic scenes~\cite{johnson2016}. With respect to quantifiers, our approach is similar to~\cite{sorodoc2016}, who use quantifiers \textit{no}, \textit{some}, and \textit{all} to quantify over sets of colored dots. Differently from ours, however, all these works tackle the issue as either a classification problem or a Visual Question Answering task, with less focus on learning the meaning representation of each cardinal/quantifier. To our knowledge, this is the first attempt to jointly investigate both mechanisms and to obtain the meaning representaton of each cardinal/quantifier as resulting from a language-to-vision mapping.

Based on their geometric intepretation, we propose to use \textbf{cosine} and \textbf{dot product} similarity between the target object and the scene as our measures for quantifiers and cardinals, respectively. The former, ranging from -1 to 1, evaluates the similarity between two vectors with respect to their orientation and irrespectively of their magnitudes. That is, the more two vectors are overall similar, the closer they are. Ideally, cosine similarity between an image depicting a \textit{dog} and a scene containing either 3 or 10 \textit{dogs} without distractors (hence, `all') should be equal to 1. Therefore, it would indicate that the proportion of `dogness' in the scene is highest. Dot product, on the other hand, is defined as the product of the cosine between two vectors and their Euclidean magnitudes. By taking into account the magnitudes, this measure ideally encodes information regarding the number of times a target object is repeated in the scene. In the above-mentioned example, indeed, dot product would be 3 and 10, respectively. In this simplified setting, thus, it would be equal to the number of \textit{dogs}.

Furthermore, we propose that the `objective' meaning of each cardinal/quantifier can be learned by means of a cross-modal mapping (see Figure~\ref{fig:diagram}) between the linguistic representation of the target object and its quantity (either exact or fuzzy) in a visual scene. To test our hypotheses, we carry out a proof-of-concept on the synthetic datasets we describe in Section~\ref{sec:data}. First, we explore our visual data by means of the two proposed similarity measures (\S~\ref{ref:onlyvision}). Second, we learn the meaning representations of cardinals and quantifiers and evaluate them in the task of retrieving unseen combinations of targets/distractors (\S~\ref{ref:cmapping}). As hypothesized, the two quantification mechanisms turn out to be better accounted for by models capitalizing on the expected similarity measures.

\section{Data}
\label{sec:data}

In order to test our hypothesis, we need a dataset of visual scenes which crucially include multiple objects. Moreover, some objects in the scene should be repeated, so that we might say, for instance, that out of 5 objects `three'/`most' are \textit{dogs}. Although a large number of image datasets are currently available (see \newcite{mscoco} among many others), no one fully satisfies these requirements. Typically, images depict one salient object and even when multiple salient objects are present, only a handful of cases contain both targets and distractors~\cite{sos2015,sos2016}. To bypass these issues, in the present work we experiment with synthetic visual scenes (hence, scenarios) that are made up by at most 9 images each representing one object. The choice of using a `patchwork' of object-depicting images is motivated by the need of representing a reasonably large variability (e.g. `few' refer to scenes containing 2 target objects out of 7 as well as 1/5, 4/9, etc.). This way, we avoid matching a quantifier always with the same number of target objects (except \textit{no}, that is always represented by 0 targets), and allow cardinals to be represented by scenes with different numbers of distractors. At the same time, we get rid of any issues related to object localization.

We experiment with quantifiers (hence, Qs) \textit{no}, \textit{few}, \textit{most}, and \textit{all}, which we defined \textit{a priori} by ratios 0\%, 1-49\%, 51-99\% and 100\%, respectively. Consistently with our goals, this arguably simplified setting does neither take into account pragmatic uses of Qs (i.e. we treat them as lying on an ordered scale) nor reflect possible overlappings. For these reasons, we avoid using quantifiers as \textit{some} whose meaning overlaps with the meaning of many others. As far as cardinals (hence, Cs) are concerned, we experiment with scenarios in which the cardinality of the targets ranges from 1 to 4. Cs up to 4 are acquired by children incrementally at subsequent stages of their development, with higher numbers being learned upon this knowledge with the ability of counting~\cite{barner2009}. Also, Cs ranging from 1 to 3-4 are widely known to exhibit some peculiar properties (i.e. their exact number can be immediately and effortlessly grasped) due to which they are usually referred to as `subitizing' range~\cite{piazza2011,railo2016}.

\begin{table}[t!]
\centering
\begin{tabular}{|cccc|cccc|}
\hline
\multicolumn{4}{|c|}{\textbf{Train-q}} & \multicolumn{4}{c|}{\textbf{Train-c}} \\ \hline
no        & few       & most      & all      & one      & two     & three     & four     \\ \hline
0/1       & 1/6       & 2/3       & 1/1      & 1/1      & 2/2     & 3/3       & 4/4      \\ 
0/2       & 2/5       & 3/4       & 2/2      & 1/3      & 2/3     & 3/4       & 4/5      \\ 
0/3       & 2/7       & 3/5       & 3/3      & 1/4      & 2/5     & 3/5       & 4/6      \\ 
0/4       & 3/8       & 4/5       & 4/4      & 1/6      & 2/7     & 3/8       & 4/7      \\ \hline
\multicolumn{4}{|c|}{\textbf{Test-q}} & \multicolumn{4}{c|}{\textbf{Test-c}} \\ \hline
no        & few       & most       & all       & one      & two      & three      & four     \\ \hline
0/5       & 1/7       & 4/6        & 5/5       & 1/2      & 2/4      & 3/7        & 4/8      \\
0/8       & 4/9       & 6/8        & 9/9       & 1/7      & 2/9      & 3/9        & 4/9      \\ \hline
\end{tabular}
\caption{Combinations in Train and Test.}\label{ref:cd}
\end{table}

\subsection{Building the scenarios}

We use images from ImageNet~\cite{imagenet}. Starting from the full list of 203 concepts and corresponding images extracted by~\newcite{cassani2014}, we discarded those concepts whose corresponding word had low/null frequency in the large corpus used in~\cite{baroni2014don}. To get rid of issues related to concept identification, we used a single representation for each of the 188 selected concepts. Technically, we computed a centroid vector by averaging the 4096-dimension visual features of the corresponding images, which were extracted from the \textit{fc7} of a CNN~\cite{simonyan2014very}. We used the VGG-19 model pretrained on the ImageNet ILSVRC data~\cite{russakovsky2015} implemented in the MatConvNet toolbox~\cite{matconvnet}. Centroid vectors were reduced to 100-d via PCA and further normalized to length 1 before being used to build the scenarios. When building the scenarios, we put the constraint that distractors have to be different from each other. Moreover, only distractors whose visual cosine similarity with respect to the target is lower than the average are selected. For each scenario, target and distractor vectors are summed together. As a result, each scenario is represented by a 100-d vector.

We also experimented with scenarios where vectors are concatenated to obtain a 900-d vector (empty `cells' are filled with 0s vectors) and further reduced to 100-d via PCA. Since the pattern of results in the only-vision evaluation (see \S~\ref{ref:onlyvision}) turned out to be similar to the results obtained in the `summed' setting, due to space limitations we will only focus on the `summed' setting.

\subsection{Datasets}\label{ref:datasets}

We built one dataset for Cs and one for Qs, each containing 4512 scenarios.\footnote{A visual representation of our scenarios is provided in the rightmost side of Figure~\ref{fig:diagram}, while Figure~\ref{fig:scenario} is only intended to provide a more intuitive overview of the task.} We then split each of the two in one 3008-datapoint Training Dataset (\textbf{Train}) for training and validation and one 1504-datapoint Testing Dataset (\textbf{Test}) for testing. The two datasets were split according to their `combinations', that is the mixture of targets and distractors in the scenario. As reported in Table~\ref{ref:cd}, we kept 4 different combinations for each C/Q in Train and 2 in Test. Note that the numerator refers to the number of targets, the denominator to the total number of objects. The number of distractors is thus given by the difference between the two values. To illustrate, in Train-q `few' is represented by scenarios 1/6, 2/5, 2/7, and 3/8, whereas in Test-q `few' is represented by scenarios 1/7 and 4/9. The initial 4512 scenarios have been obtained by building a total of 24 different scenarios (6 combinations * 4 C/Q classes) for each of the 188 objects. A particular effort has been paid in making the datasets as balanced as possible. When designing the combinations for `few' and `most', for example, we controlled for the proportion of targets in the scene, in order to avoid making one of the two easier to learn. Also, combinations were thought to avoid biasing cardinals toward fixed proportions of targets/distractors.

\begin{figure}[t!]
\begin{center}
\hfill
\subfigure{\includegraphics[width=0.49\linewidth]{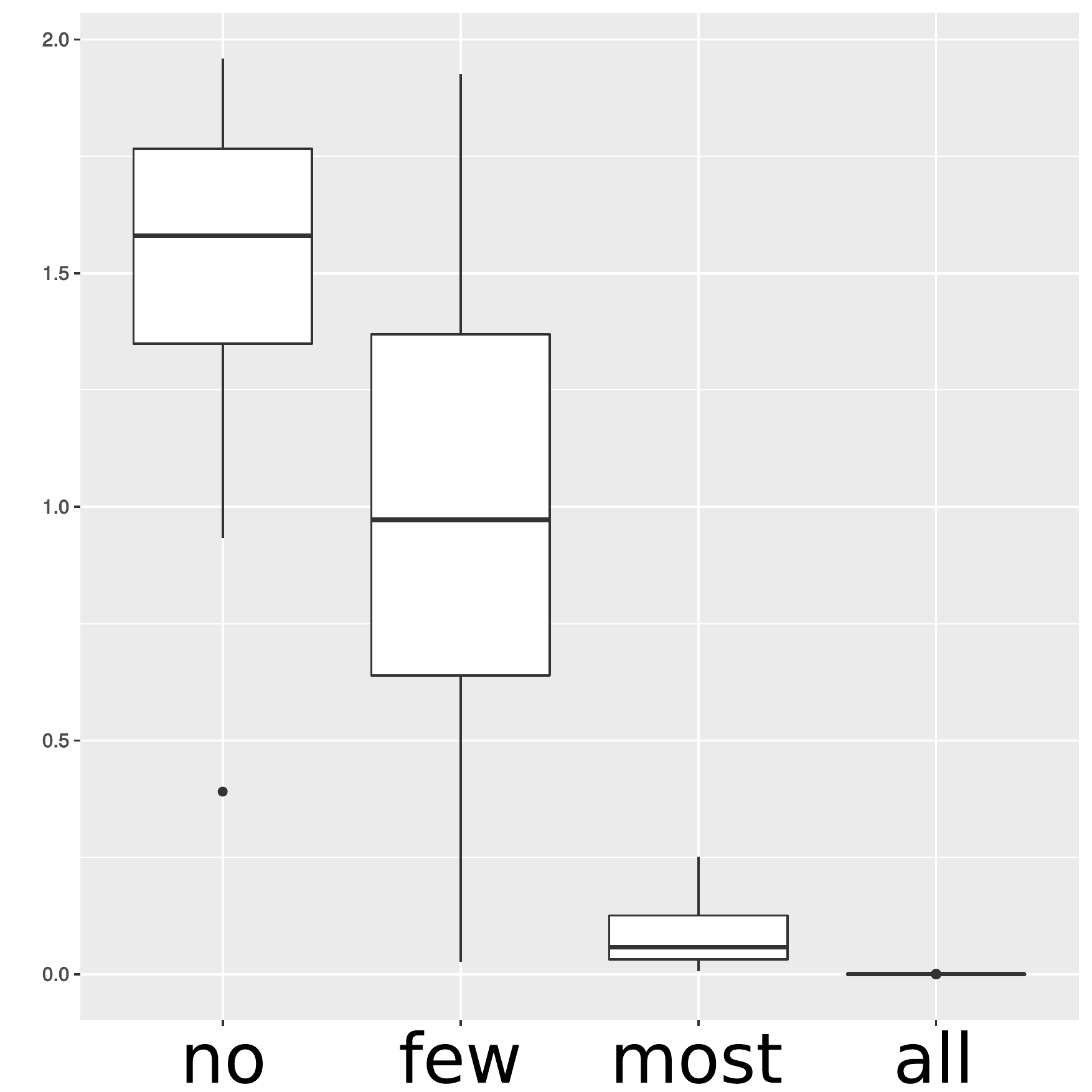}}
\hfill
\subfigure{\includegraphics[width=0.49\linewidth]{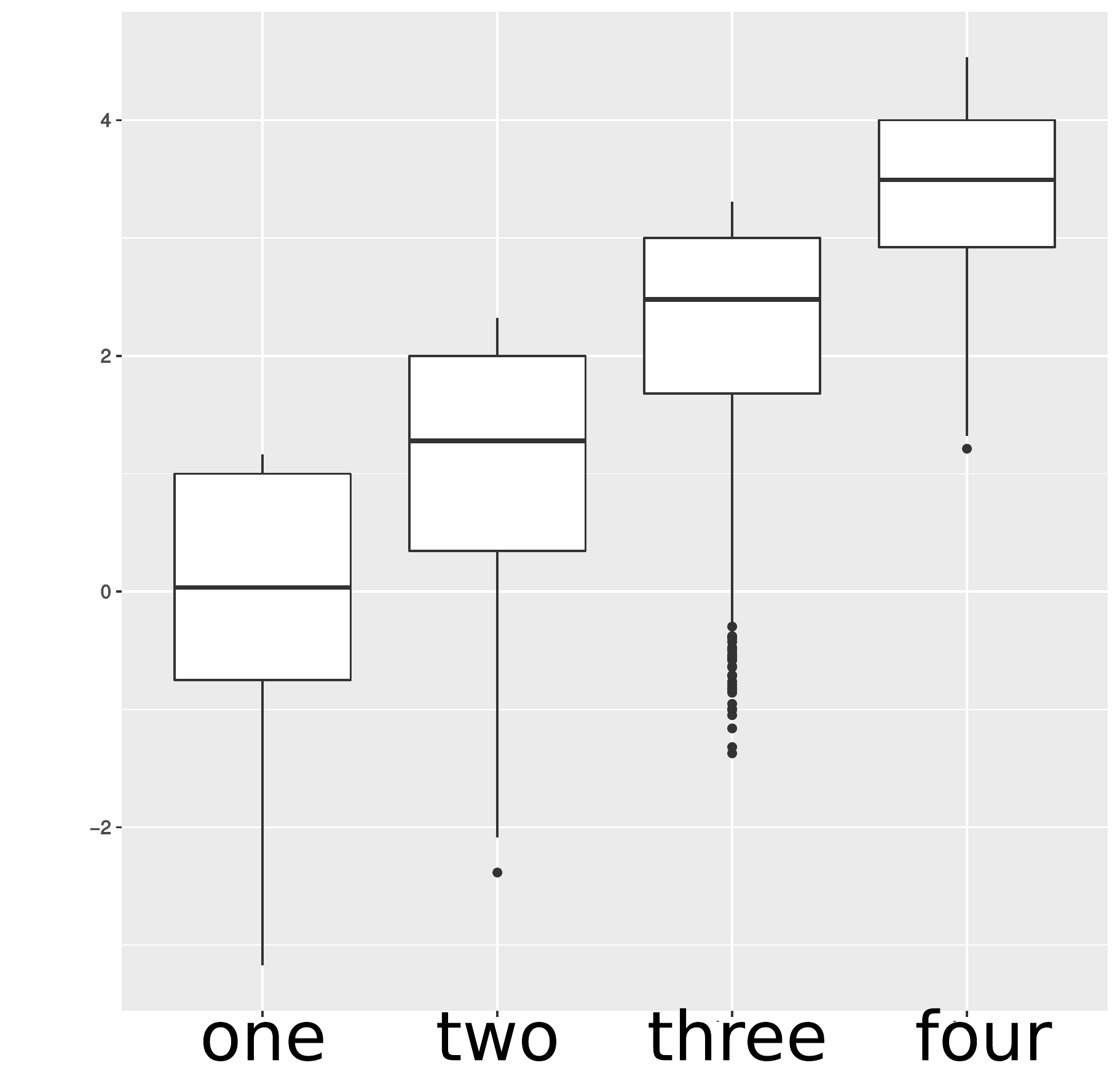}}
\hfill
\end{center}
\caption{Left: quantifiers against cosine distance. Right: cardinals against dot product.}\label{ref:box}
\end{figure}

\begin{figure}[t!]
\begin{center}
\hfill
\subfigure{\includegraphics[width=0.49\linewidth]{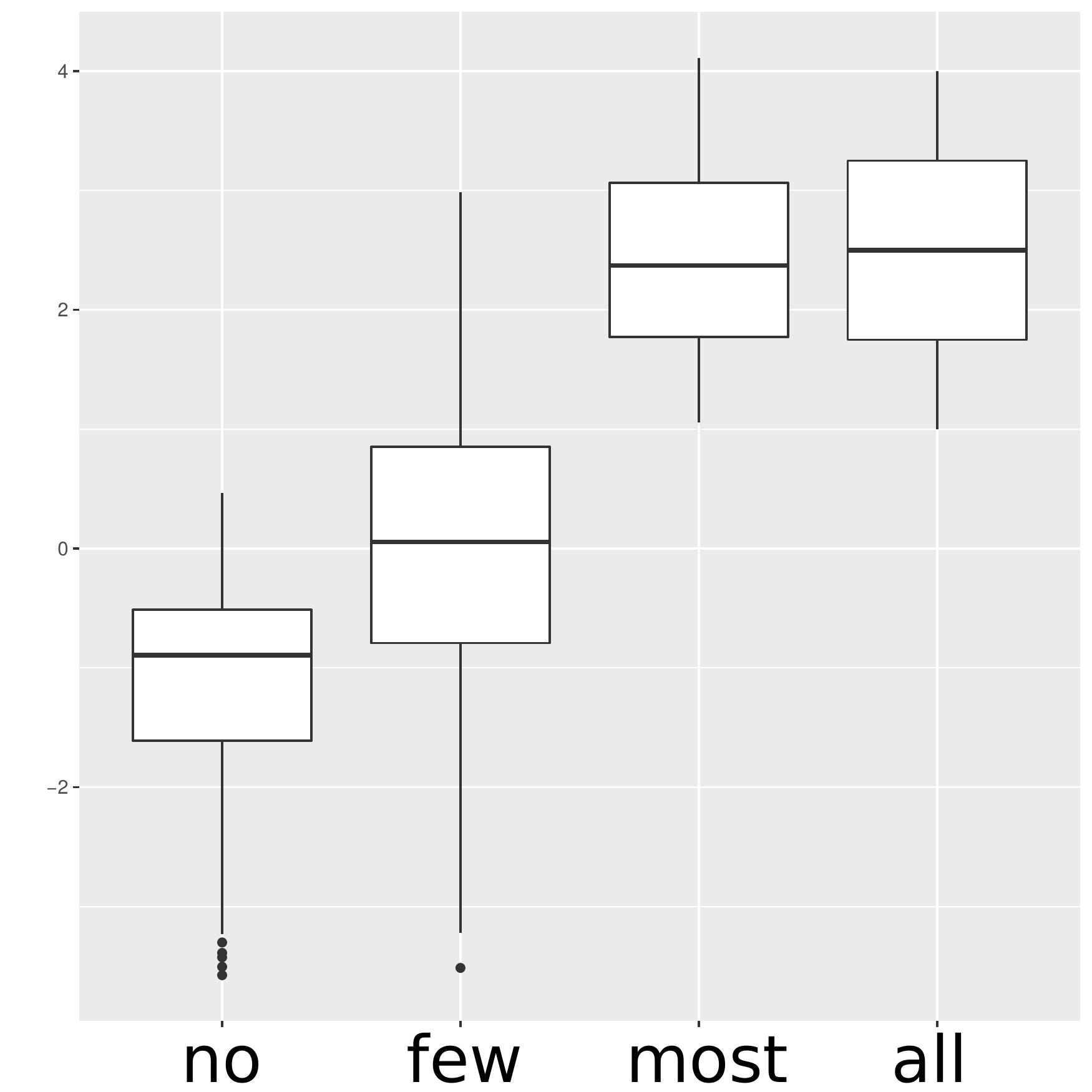}}
\hfill
\subfigure{\includegraphics[width=0.49\linewidth]{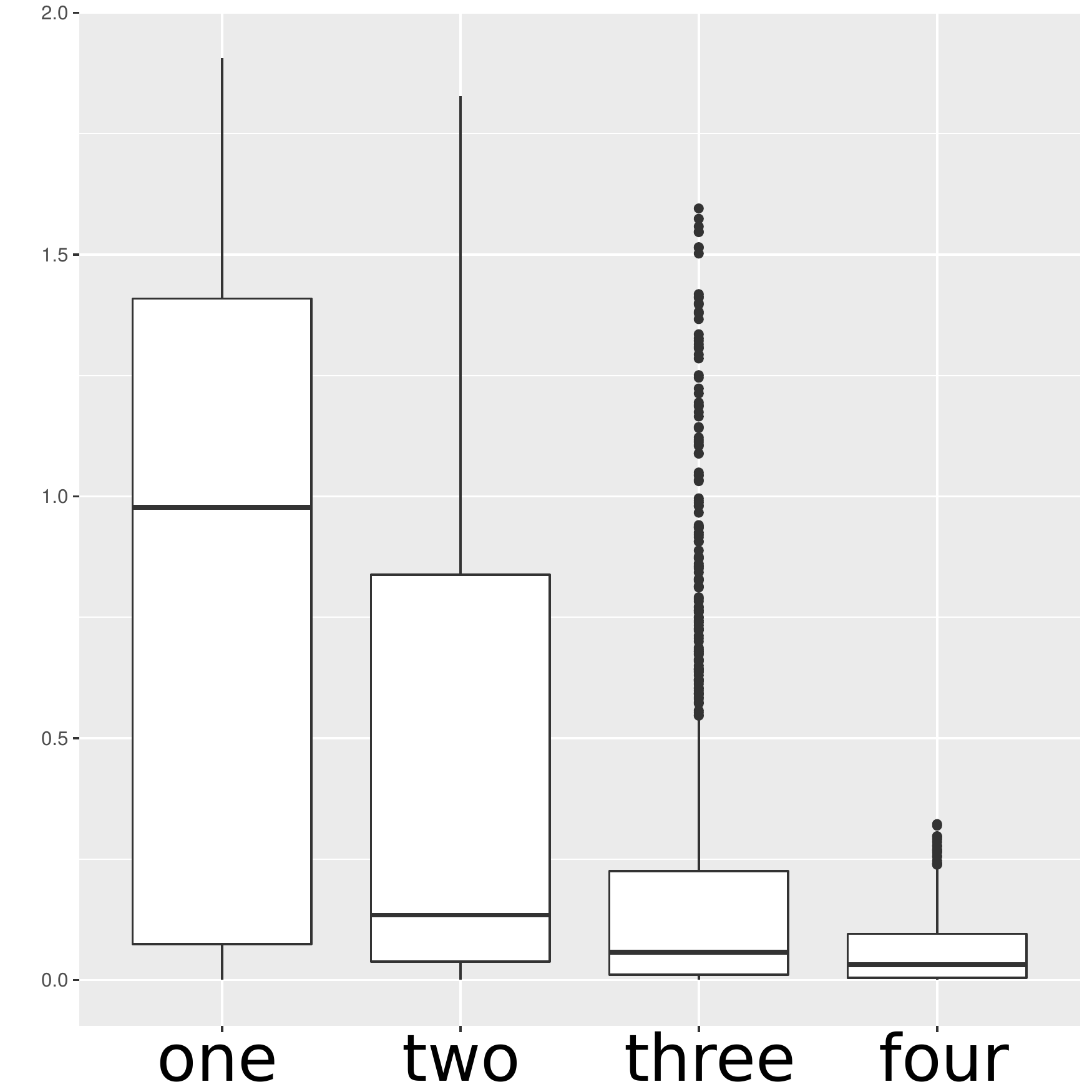}}
\hfill
\end{center}
\caption{Left: quantifiers against dot product. Right: cardinals against cosine distance.}\label{ref:box2}
\end{figure}

\section{Experiments}
\label{sec:exp}

\subsection{Only-vision evaluation}\label{ref:onlyvision}

As a first step, we carry out a preliminary evaluation aimed at exploring our visual data. If our intuition about the information encoded by the two similarity measures is correct (see \S~\ref{sec:intro}), we should observe that cosine is more effective than dot product in distinguishing between different Qs, while the latter should be better than cosine for Cs. Moreover, Qs/Cs should lie on an ordered scale. To test our hypothesis, we compute cosine distances (i.e. 1${-}$cosine, to avoid negative values) and dot product similarity for each target-scenario pair in both Train and Test (e.g. \textit{dog} vs 2/5 \textit{dogs}). Figure~\ref{ref:box} reports the distribution of Qs with respect to cosine (left) and Cs with respect to dot product (right) in Train. As can be seen from the boxplots, both Qs and Cs are ordered on a scale. In particular, cosine distance is highest in \textit{no} scenarios (where the target is not present), lowest in \textit{all} scenarios. For Cs, dot product is highest in \textit{four} scenarios, lowest in \textit{one} scenarios.

Our intuition is further confirmed by the results of a radial-kernel SVM classifier fed with either cosine or dot product similarities as predictors.\footnote{We experimented with linear, polynomial, and radial kernels. We only report results obtained with default radial kernel, that turned out to be the overall best model.} Qs are better predicted by cosine than dot product (78.6\% vs 63.8\%), whereas dot product is a better predictor of Cs than cosine (68.7\% vs 44.7\%). As shown in Figure~\ref{ref:box2}, the ordered scale is indeed represented to a much lesser extent when Qs are plotted against dot product (left) and Cs against cosine (right). A similar pattern of SVM results and similar plots emerged when experimenting with Test.

\begin{figure}[t!]
\begin{center}
  \includegraphics[width=0.7\linewidth]{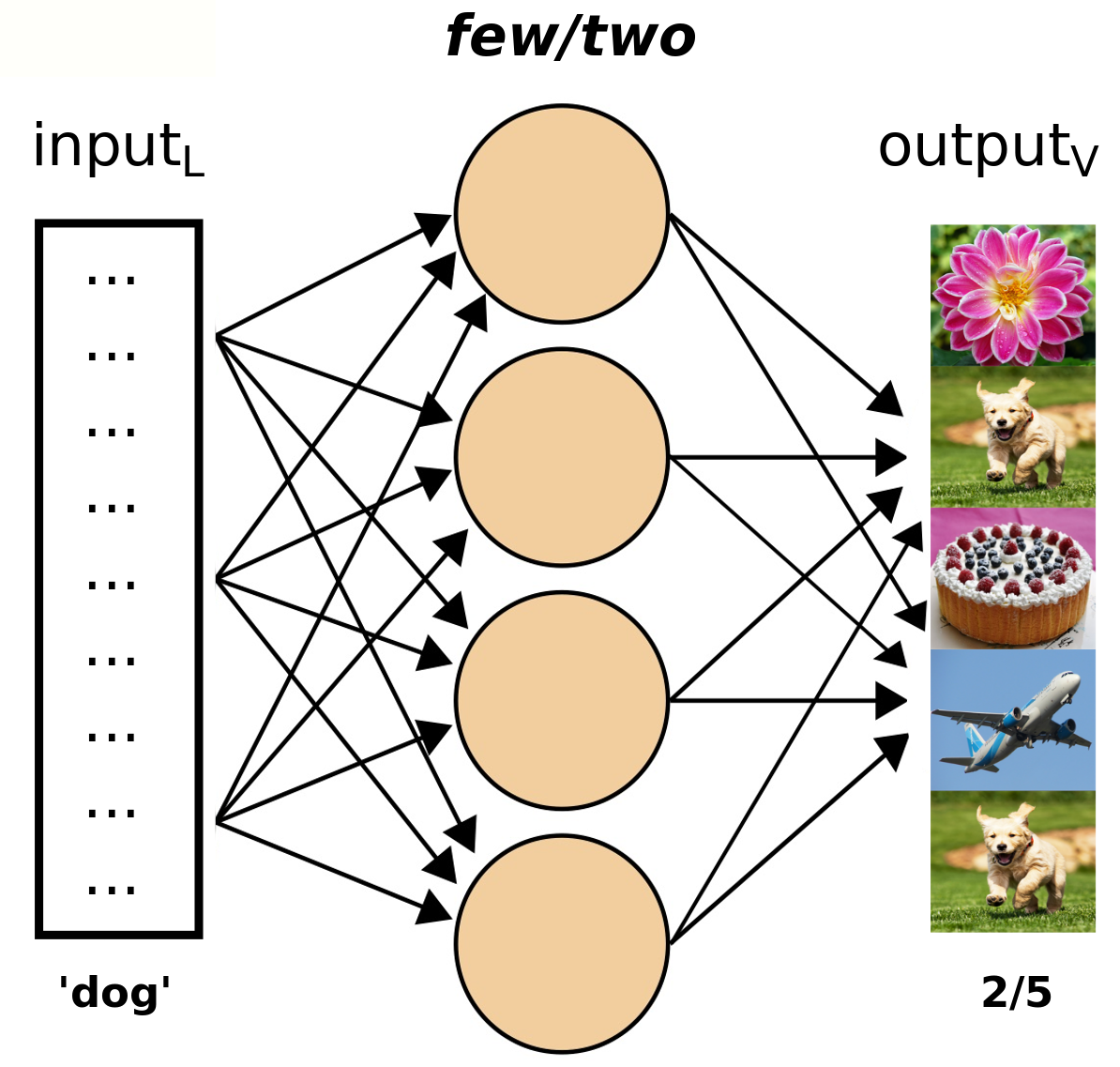}
\end{center}
\caption{One learning event of our proposed cross-modal mapping. Cosine is used for quantifiers (\textit{few}), dot product for cardinals (\textit{two}).}\label{fig:diagram}
\end{figure}

\subsection{Cross-modal mapping}\label{ref:cmapping}

Our core proposal is that the meaning of each C/Q can be learned by means of a cross-modal mapping between the linguistic representation of the target object (e.g. \textit{dog}, \textit{mug}, etc.) and a number of scenarios representing the target object in a given C/Q setting (e.g. `two'/`few' \textit{dogs}). In our approach, each word (e.g. \textit{dog}) is represented by a 400-d embedding built with the \texttt{CBOW} architecture of \texttt{word2vec}~\cite{mikolov2013} and the best-predictive parameters of~\newcite{baroni2014don} on a 2.8B tokens corpus. The original 400-d vectors are further reduced to 100-d via PCA before being fed into the model.

Figure~\ref{fig:diagram} reports a single learning event of our proposed model. Each C/Q (e.g. \textit{two}, \textit{few}) is learned as a separate function that maps each of the 188 words representing our selected concepts to its corresponding 4 scenarios in Train (see \S~\ref{ref:datasets}). To illustrate, the meaning of \textit{few} is learned by mapping each word into the 4 visual scenes where the amount of `targetness' is less than 50\% (see \S~\ref{sec:data}), whereas \textit{two} is learned by mapping each word to the scenarios where the number of targets is 2, and so on. This mapping, we conjecture, would mimic the multimodal mechanism by which children acquire the meaning of both Cs and Qs (see~\newcite{halberda2008}). Once learned, the function representing each C/Q can be evaluated against scenarios containing an unseen mixture of (known) target objects and distractors. If it has encoded the correct meaning of the quantified expression, the function will retrieve the unseen scenarios containing the correct quantity (either exact or fuzzy) of target objects.

\begin{table}[t]
\centering
\begin{tabular}{|l|ll|ll|ll|}
\hline
      & \multicolumn{2}{c|}{\textbf{lin}}                                    & \multicolumn{2}{c|}{\textbf{nn-cos}}                                 & \multicolumn{2}{c|}{\textbf{nn-dot}}                                 \\ \cline{2-7} 
      & \multicolumn{1}{c}{\textit{mAP}} & \multicolumn{1}{c|}{\textit{P2}} & \multicolumn{1}{c}{\textit{mAP}} & \multicolumn{1}{c|}{\textit{P2}} & \multicolumn{1}{c}{\textit{mAP}} & \multicolumn{1}{c|}{\textit{P2}} \\ \hline
no    & 0.78                             & 0.65                              & \textbf{0.87}                    & {\ul 0.77}                        & 0.54                             & 0.37                              \\
few   & 0.59                             & 0.39                              & \textbf{0.68}                    & {\ul 0.51}                        & 0.59                             & 0.43                              \\
most  & 0.61                             & 0.36                              & 0.60                             & 0.29                              & \textbf{0.62}                    & {\ul 0.45}                        \\
all   & 0.75                             & 0.66                              & \textbf{1}                       & {\ul 1}                           & 0.33                             & 0.12                              \\ \hline
one   & 0.44                             & 0.30                              & 0.38                             & 0.21                              & \textbf{0.61}                    & {\ul 0.45}                        \\
two   & 0.35                             & 0.15                              & 0.38                             & 0.21                             & \textbf{0.57}                    & {\ul 0.43}                        \\
three & 0.38                             & 0.16                              & 0.36                             & 0.13                              & \textbf{0.56}                    & {\ul 0.40}                        \\
four  & 0.65                             & 0.47                              & 0.75                    & 0.60                              & \textbf{0.76}                             & {\ul 0.61}                        \\ \hline
\end{tabular}
\caption{R-target. \textit{mAP} and \textit{P2} for each model.}\label{ref:targettab}
\end{table}

We experiment with three different models: linear (\textbf{lin}), cosine neural network (\textbf{nn-cos}), dot-product neural network (\textbf{nn-dot}). The first model is a simple linear mapping. The second is a single-layer neural network (activation function ReLU) that maximizes the cosine similarity between input (linguistic) and output vector (visual). The third is a similar neural network that approximates to 1 the dot product between input and output. We evaluate the mapping functions by means of a retrieval task aimed at picking up the correct scenarios from Test among the set of 8 scenarios built upon the same target object. Recall that in Test there are 2 combinations * 4 C/Q classes for each concept.

\textbf{Results} As reported in Table~\ref{ref:targettab}, nn-cos is overall the best model for Qs, whereas nn-dot is the best model for Cs. In particular, mean average precision (\textit{mAP}) is higher in nn-cos for 3 out of 4 Qs, with only \textit{most} reaching slightly better mAP in Q nn-dot due to the high number of cases confounded with \textit{all} by the Q nn-cos model (see Table~\ref{ref:qnncos}). Conversely, both mAP and precision at top-2 positions (\textit{P2}) for Cs are always higher in nn-dot compared to the other models. From a qualitative analysis of the results, it emerges that both the best-predictive models make `plausible' errors, i.e. they confound Cs/Qs that are close to each other in the ordered scale. Table~\ref{ref:qnncos} reports the confusion matrices for the best performing models. Besides retrieving more cases of \textit{all} instead of (correct) \textit{most}, the Q nn-cos model often confounds \textit{few} with \textit{no}. Similarly, the C nn-dot model often confounds \textit{three} with \textit{four}, \textit{one} with \textit{two}, \textit{two} with \textit{three}, and so on. Overall, both models pick up very few or no responses that are on the opposite end of the `scale', thus suggesting that the meaning representation they learn encodes, to a certain extent, information about the ordered position of the quantified expressions.

\begin{table}[]
\centering
\begin{tabular}{|l|cccc|}
\hline
              & no  & few & most & all \\ \hline
no   & \textbf{288} & 88           & 0             & 0            \\
few  & 141          & \textbf{191} & 38            & 6            \\
most & 0            & 0            & 111           & {\ul 265}    \\
all  & 0            & 0            & 0             & \textbf{376} \\ \hline
               & one & two & three & four \\ \hline
one   & \textbf{168} & 113          & 54             & 41            \\
two   & 64           & \textbf{136} & 124            & 52            \\
three & 23           & 80           & 130            & {\ul 145}     \\
four  & 10           & 24           & 72             & \textbf{272}  \\ \hline

\end{tabular}
\caption{Top: Q nn-cos, number of cases retrieved in top-2 positions. Bottom: same for C nn-dot.}\label{ref:qnncos}
\end{table}

\section{Discussion}
\label{sec:disc}

We propose that the meaning of Cs and Qs can be learned by means of a language-to-vision mapping, and we show that two models capitalizing on dot product and cosine better account for Cs and Qs, respectively. In future research, we plan to further investigate this issue by using real-scene images to avoid constraining the visual data. Moreover, we plan to experiment with a broader set of quantifiers (e.g. \textit{some}, \textit{almost all}, etc.) and higher cardinals. The latter investigation, in particular, would allow us to verify whether our approach is suitable for the (potentially infinite) set of `cardinal functions' beyond the subitizing range. If so, we might observe that the models keep making cognitively plausible errors, picking items that are close to the target one in the ordered scale. This evidence, we believe, would further motivate our `one quantifed expression, one function' approach, which is partially inspired by the evidence that, in human brain, so-called number neurons are tuned to preferred numbers~\cite{nieder2016}. Simplifying somewhat, each number would activate specific neurons. Finally, we believe that taking into account speakers' uses of Cs and Qs would constitute the natural next step toward a complete modelling of the meaning of quantified expressions.

\section*{Acknowledgments}
We are very grateful to Germ\'{a}n Kruszewski for the invaluable contribution in developing and discussing the intuitions behind this work. We are also grateful to Marco Baroni, Aur\'elie Herbelot, Gemma Boleda and Ravi Shekhar for their advice and feedback. We gratefully acknowledge the support of NVIDIA Corporation with the donation of the GPUs used in our research, and the iV\&L Net (ICT COST Action IC1307) for funding the second author's research visit aimed at working on this project.

\bibliography{paper}
\bibliographystyle{eacl2017}

\end{document}